\documentclass[sigconf,nonacm]{acmart}

\usepackage{multirow}
\usepackage{tikz}
\usepackage{booktabs}
\usepackage{graphicx}
\usepackage{amsmath}
\usepackage{longtable}
\usepackage{subcaption}
\usepackage{enumitem}
\setlist{nolistsep}
\usepackage{balance}
\usepackage{microtype}
\usepackage{microtype}

\begin{document}

\title{DRGW: Learning Disentangled Representations for Robust Graph Watermarking}

\author{Jiasen Li}
\affiliation{%
  \institution{Institute of Information Engineering, Chinese Academy of Sciences}
  \institution{School of Cyber Security, University of Chinese Academy of Sciences}
  \institution{State Key Laboratory of Cyberspace Security Defense}
  \city{Beijing}
  \country{China}
}
\email{lijiasen@iie.ac.cn}

\author{Yanwei Liu}
\authornote{Corresponding author\\[8pt]
\textcolor{red}{\textbf{Proceedings of the ACM Web Conference 2026 (WWW ’26), April 13-17, 2026, Dubai, United Arab Emirates}}}
\affiliation{%
  \institution{Institute of Information Engineering, Chinese Academy of Sciences}
  \institution{State Key Laboratory of Cyberspace Security Defense}
  \city{Beijing}
  \country{China}
}
\email{liuyanwei@iie.ac.cn}

\author{Zhuoyi Shang}
\affiliation{%
  \institution{Institute of Information Engineering, Chinese Academy of Sciences}
  \institution{School of Cyber Security, University of Chinese Academy of Sciences}
  \institution{State Key Laboratory of Cyberspace Security Defense}
  \city{Beijing}
  \country{China}
}
\email{shangzhuoyi@iie.ac.cn}

\author{Xiaoyan Gu}
\authornotemark[1]
\affiliation{%
  \institution{Institute of Information Engineering, Chinese Academy of Sciences}
  \institution{School of Cyber Security, University of Chinese Academy of Sciences}
  \institution{State Key Laboratory of Cyberspace Security Defense}
  \city{Beijing}
  \country{China}
}
\email{guxiaoyan@iie.ac.cn}

\author{Weiping Wang}
\affiliation{%
  \institution{Institute of Information Engineering, Chinese Academy of Sciences}
  \city{Beijing}
  \country{China}
}
\email{wangweiping@iie.ac.cn}

\renewcommand{\shortauthors}{Jiasen Li et al.}

\begin{abstract}
Graph-structured data is foundational to numerous web applications, and watermarking is crucial for protecting their intellectual property and ensuring data provenance. Existing watermarking methods primarily operate on graph structures or entangled graph representations, which compromise the transparency and robustness of watermarks due to the information coupling in representing graphs and uncontrollable discretization in transforming continuous numerical representations into graph structures.  This motivates us to propose DRGW, the first graph watermarking framework that addresses these issues through disentangled representation learning. Specifically, we design an adversarially trained encoder that learns an invariant structural representation against diverse perturbations and derives a statistically independent watermark carrier, ensuring both robustness and transparency of watermarks. Meanwhile, we devise a graph-aware invertible neural network to provide a lossless channel for watermark embedding and extraction, guaranteeing high detectability and transparency of watermarks. Additionally, we develop a structure-aware editor that resolves the issue of latent modifications into discrete graph edits, ensuring robustness against structural perturbations. Experiments on diverse benchmark datasets demonstrate the superior effectiveness of DRGW. 
\end{abstract}


\maketitle

\section{Introduction}

The proliferation of Generative AI has accelerated data creation and dissemination across the web\cite{team2024gemini, bai2025principled}, spanning unstructured content like text\cite{nie2025llmtextsurvey} and images\cite{li2025diffusion3dsurvey} to structured data like tables\cite{borisov2025tabulargensurvey}, time-series\cite{zhang2025timeseriesmllmsurvey}, and graphs\cite{zhu2022graphgensurvey, zhang2025graphoodsurvey}. Among these, graph data is a cornerstone of the modern web, uniquely modeling complex relationships\cite{ju2024deepgrlsurvey,papadopoulos2024gnnsurvey} and hierarchies\cite{pan2024unifyingllmkg} for critical applications like social network analysis\cite{sharma2024gnnrecsyssurvey}, recommendation systems\cite{sharma2024gnnrecsyssurvey}, and knowledge discovery\cite{pan2024unifyingllmkg}. Graph representation learning\cite{ju2024deepgrlsurvey} has become a foundational paradigm, where embeddings encode structural patterns for downstream tasks like node classification\cite{kipf2017gcn,ju2024deepgrlsurvey} and link prediction\cite{li2023linkpredictionbenchmark}. This, in turn, drives widespread sharing of graph data across organizations\cite{pglobal2024cyberrisk,guan2025stsa}, exposing them to significant security risks

However, the need for data sharing conflicts with data protection. Malicious actors can leak proprietary information~\cite{ harvardprivacyattacksurvey2023}, perturb graph structures to compromise integrity~\cite{zhang2025llm4rgnn}, or exploit data for unauthorized commercial use~\cite{purpose2025}. Such breaches erode trust in data-sharing and allow intellectual property infringement to go unpunished, due to the inability to attribute the data leak's source. Collectively, these threats highlight a critical challenge: the lack of a reliable mechanism for robust \textit{provenance tracking}, \textit{integrity verification}, and \textit{copyright protection} of shared graphs that preserves their utility.

\begin{figure}[htbp]
\centering
\includegraphics[width=\columnwidth]{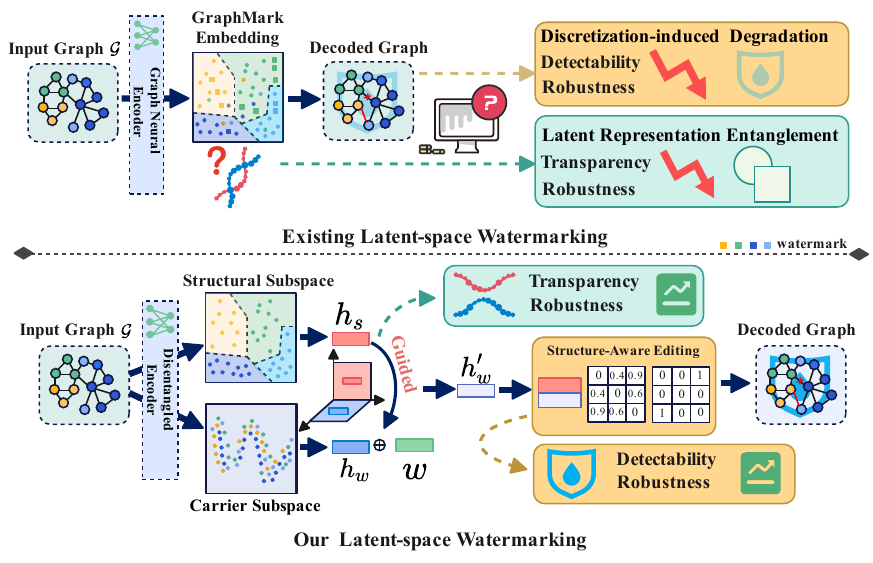}
\vspace{-2.5em}
\caption{Conceptual comparison of our method with existing scheme. The prevailing latent-space watermarking is constrained by two fundamental flaws: information entanglement, which compromises transparency and robustness, and discretization-induced watermark degradation, where the faint signal is discarded during conversion from continuous latent representation to a discrete graph structure. Our DRGW framework resolves these issues through representation disentanglement and structure-aware editing.}
\label{fig:conceptual_overview}
\vspace{-1.5em}
\end{figure}

Digital watermarking, a technique proven to be effective for copyright protection and provenance tracking in domains like multimedia~\cite{bhattacharya2021multimediawatermarkingsurvey,chen2025plugmark} and text~\cite{liu2024textwatermarkingsurvey,shang2026attestingmodellineageconsisted}, presents a
promising avenue to address this challenge. However, the direct application of these off-the-shelf algorithms to graph data faces fundamental obstacles, primarily due to its inherent properties of permutation invariance and the lack of a canonical data order~\cite{hamilton2020grlbook}. This discrepancy necessitates the development of entirely new, graph-native watermarking paradigms. An ideal graph-native watermarking system must synergistically achieve three core objectives \cite{peng2025kgmark}: \textit{Robustness} against structural modifications and adversarial attacks, \textit{Detectability} ensuring reliable watermark extraction, and \textit{Transparency} by preserving both the graph's structural integrity and its functional utility for downstream tasks. Current research on graph watermarking can be broadly categorized into two paradigms striving towards these goals.

\textbf{Structure-space watermarking}~\cite{zhao2015towards,Models} operates by directly modifying the discrete graph topology, thereby intertwining the watermark signal with the host graph's native structure. This direct coupling, however, renders the watermark fragile and highly dependent on specific local substructures, making it vulnerable to even minor perturbations and limiting its general applicability across diverse graph types.

In an attempt to overcome these limitations, \textbf{latent-space watermarking}~\cite{peng2025kgmark} shifts the embedding process to continuous representation spaces. While this mitigates dependency on a specific topology, it merely relocates the entanglement problem from the structural domain to the latent domain, as illustrated in Figure~\ref{fig:conceptual_overview}. Imposing watermarks on holistic graph embeddings—where structural, semantic, and now watermark information are non-linearly coupled—inherently compromises the transparency and robustness of the watermark. The decoder, tasked with reconstructing the discrete graph from this entangled latent representation, cannot reliably distinguish the faint watermark signal, leading to its attenuation during the discretization in transforming continuous numerical representations into adjacent matrix. Furthermore, the generative models (e.g., VAEs) often employed in this paradigm possess an inherent information bottleneck that may actively discard the watermark as noise during reconstruction, fundamentally compromising detectability and robustness.

To fundamentally overcome the intertwined challenges of information entanglement and discretization-induced watermark degradation, we propose DRGW, a disentanglement-based watermarking framework built upon three synergistic innovations. First, we design a robust graph encoder that is adversarially trained against diverse structural perturbations—including node/edge modifications and isomorphism attacks—to extract an invariant structural representation ($h_s$) while deriving a statistically independent watermark carrier ($h_w$) through mutual information minimization. This principled separation inherently ensures both transparency and robustness by isolating watermark operations from the graph's core functional properties. Second, we introduce a Graph-aware Invertible Neural Network (INN) that performs lossless, topology-conditioned transformations on the watermark carrier, confining perturbations to a minimal subspace to enhance watermark detectability while preserving transparency. Finally, we develop a structure-aware editor that 
addresses the discretization loss problem by translating continuous watermark signals into discrete graph edits using the preserved structural knowledge from $h_s$, ensuring robust watermark detectability against quantization errors.

The main contributions of this work are summarized as follows:
\begin{itemize}
  \item We propose DRGW, a latent-space watermarking framework that explicitly reduces the fragility of watermarks arising from the entanglement of information, via learning the invariant structural representations of graphs while deriving an orthogonal watermark space.
  \item We devise a graph-aware INN for high-fidelity watermark transformation and a structure-aware editor for alleviating the discretization-induced watermark degradation.
  
  \item Extensive experiments demonstrate that DRGW achieves the state-of-the-art performance, exhibiting exceptional robustness against diverse attacks while maintaining superior structural and functional fidelity.
\end{itemize}

\section{Related Work}
\subsection{Protecting Privacy in Graphs}
Prior work on graph data security has primarily focused on privacy preservation, exploring several distinct paradigms. For instance, some methods modify the graph's structure directly; these include graph anonymization~\cite{hoang2024personalized}, and randomization~\cite{ying2008randomizingsocialnetworks}. 
Another line of work focuses on differentially private data generation \cite{dwork2014dpfoundations} or model-based defense against membership inference attacks \cite{chen2025mpgstack}.A common drawback of these techniques is that in preventing identity and attribute disclosure, they inevitably alter the graph's structure and reduce its utility. In contrast, watermarking seeks to enable provenance tracking without sacrificing data utility, complementing existing graph security solutions.

\subsection{Graph Watermarking}
Existing graph watermarking methods can be categorized by their operational domain:
\textbf{Structure-space watermarking }~\cite{zhao2015towards,Models} operates directly on graph topology. Seminal work by Zhao et al.~\cite{zhao2015towards} embeds watermarks by modifying specific subgraphs identified through node structure descriptors. While providing theoretical uniqueness guarantees, these methods suffer from fragility against structural perturbations and limited applicability across diverse graph types due to their dependency on local topological features.
\textbf{Latent-space watermarking} represents a paradigm shift to continuous representations. KGMark~\cite{peng2025kgmark} employs diffusion models to embed watermarks in knowledge graph embeddings, while other approaches leverage variational autoencoders ~\cite{simonovsky2018graphvae} and generative adversarial networks ~\cite{decao2018molgan} for similar purposes. Although these methods improve generalizability, they introduce new challenges in information entanglement and discretization reconstruction, as analyzed in the Introduction.

\subsection{Representation Disentanglement}
Recent advances have explored separating graph representations into interpretable factors \cite{ma2025gomkcn}, typically for tasks like explainable AI and domain adaptation. Techniques include mutual information minimization \cite{sanchez2019disentanglemi}, adversarial training \cite{zheng2024adgcn, kuroda2022adversarialdisentangle}, and orthogonal subspace projection \cite{lui2022orthogonalgnn}. However, these methods have not been exploited in watermarking context. Our work is the first to leverage representation disentanglement specifically for resolving the fundamental challenges in graph watermarking, creating a new intersection between disentangled representation learning and digital watermarking.

\section{Threat Model}
\label{sec:threat_model}
We define our threat model to specify the roles of the data owner and the adversary in a realistic data-sharing context.

\textbf{The Data Owner (Defender).}
The owner's goal is to reliably verify their ownership of a suspicious graph $\mathcal{G}'_w$ possibly attacked by an adversary. To do so, they operate in two distinct phases. During the \textit{Embedding Phase}, 
they have full access to their original graph $\mathcal{G}$ to embed a secret watermark $w$. During the \textit{Verification Phase}, they use the watermark $w$ and our public verification framework to test the suspicious graph and formally prove ownership.

\textbf{The Adversary.}
The adversary's goal is to evade ownership detection by removing or corrupting the embedded watermark $w$ from the graph $\mathcal{G}_w$, creating a modified version $\mathcal{G}'_w$ that still preserves a high degree of structural and functional utility. We assume a strong adversary with the following capabilities:
\begin{itemize}[leftmargin=*, nosep]
    \item \textbf{Full Knowledge of the Graph:} The adversary possesses the complete watermarked graph $\mathcal{G}_w$, including its nodes, edges, and any associated features.
    \item \textbf{Knowledge of the Algorithm:} In line with Kerckhoffs's Principle \cite{Kerchhoffs}, we assume the adversary may know the watermarking algorithm, but not the owner's watermark $w$.
    \item \textbf{Graph Manipulation:} The adversary can employ a variety of attacks. Our model is designed to be robust against structural perturbations (e.g., node/edge addition and deletion), isomorphism attacks (node relabeling), and sophisticated adaptive attack designed to degrade watermark detectability.
\end{itemize}

\section{Method}
To handle the challenges caused by information entanglement and discretization graph reconstruction, we propose DRGW. It embeds a digital watermark $w$ into a given graph $\mathcal{G}=(V, E)$ by leveraging three synergistic innovations: a disentangled encoder (Sec.~\ref{sec:encoder}) that separates the graph into invariant structural ($h_s$) and independent watermark ($h_w$) representations to ensure watermark transparency and robustness; a Graph-aware INN (Sec.~\ref{sec:inn}) that performs lossless transformations on the carrier to enhance detectability while preserving transparency; and a structure-aware editor (Sec.~\ref{sec:editor}) that resolves the \textit{discretization loss problem} by translating continuous signals into robust discrete graph edits.

\subsection{Overview}
\begin{figure*}[t!]
  \centering
  \includegraphics[width=0.95\textwidth]{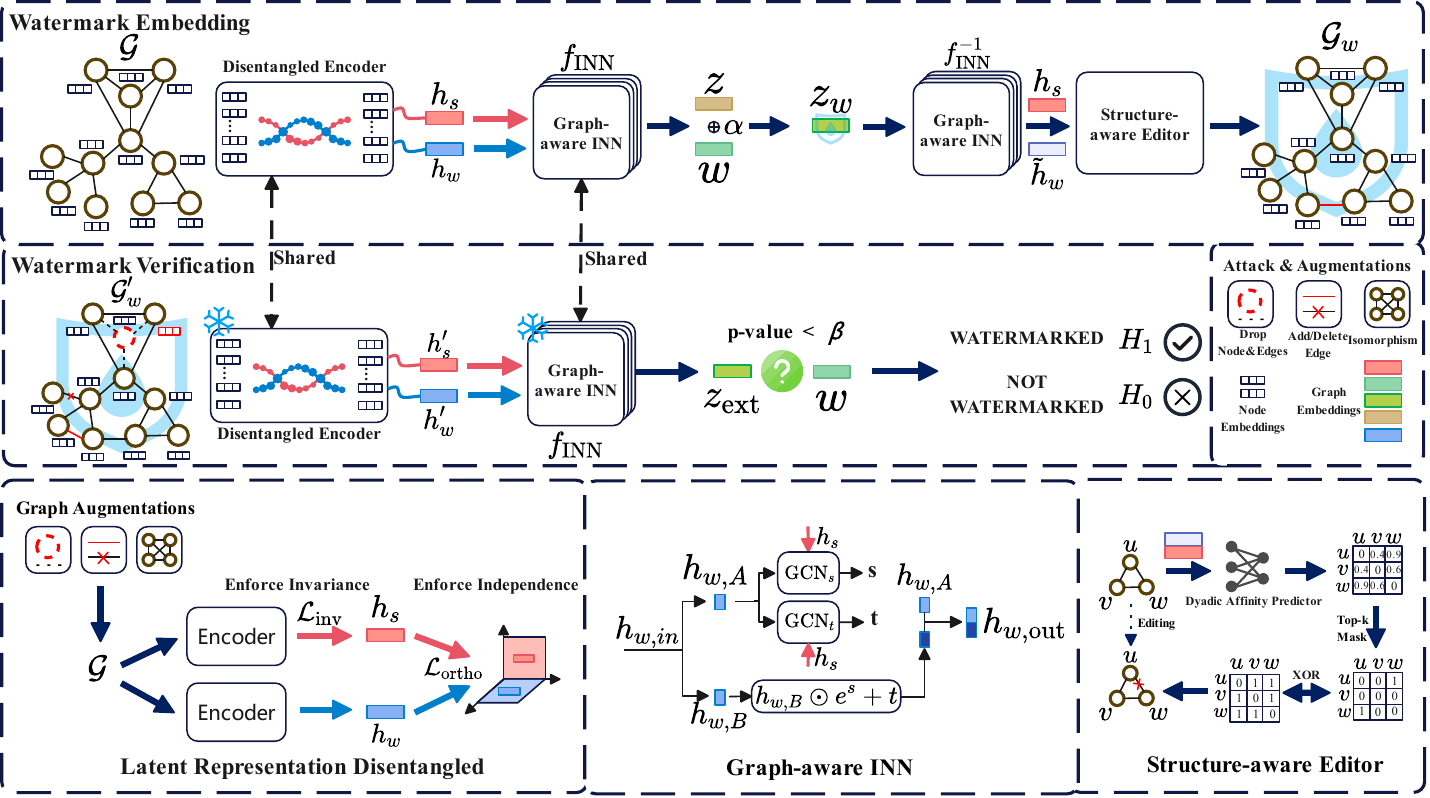}
  \vspace{-1em}
  \caption{Architecture of the DRGW. The top in the figure illustrates the watermark embedding and verification pipelines. The bottom in the figure details the mechanisms of our key components: the Disentangled Encoder (left) uses contrastive and orthogonality losses to separate representations; the Graph-aware INN (middle) conditions its transformation on the structural representation $h_s$; and the structure-aware Editor (right) uses the learned representations to predict robust graph edits.}
  \label{fig:framework_overview}
  \vspace{-1em}
\end{figure*}

As illustrated in Figure~\ref{fig:framework_overview}, the DRGW framework operates through two sequential stages: watermark embedding and verification. In the embedding stage, an encoder $\Phi_E$ disentangles an input graph $\mathcal{G}$ into a structural component $h_s$ and a carrier $h_w$. The carrier is mapped by an INN, $f_{\text{INN}}$, into a latent space where the watermark $w$ is injected. The Graph-aware INN's inverse, $f_{\text{INN}}^{-1}$, maps this back to a modified carrier, which a structure-aware editor $\Phi_D$ uses to generate a discrete edit plan $\Delta E$, producing the watermarked graph $\mathcal{G}_w$. The verification stage uses the same encoder $\Phi_E$ and INN $f_{\text{INN}}$, with their parameters frozen and publicly shared. A suspect graph is first encoded to obtain its carrier representation, which is then transformed by the Graph-aware INN to extract a potential watermark. Finally, this extracted signal is statistically compared against the original $w$ via a likelihood-based hypothesis test to confirm the watermark's presence.

\subsection{Watermark Embedding}
\label{sec:embedding}

The watermark embedding process transforms an input graph $\mathcal{G}$ into a watermarked version $\mathcal{G}_w$. We formulate this as a constrained optimization problem. Given a host graph $\mathcal{G} = (V, E)$ and a watermark $w \sim \mathcal{N}(0, I)$, where $I$ is the identity matrix, we seek an embedding function $\mathcal{E}$ that satisfies:

\begin{equation}
\begin{aligned}
& \underset{\mathcal{E}}{\text{maximize}} & & \mathbb{P}[\mathcal{D}(\mathcal{A}(\mathcal{E}(\mathcal{G}, w))) = w] \\
& \text{subject to} & & \text{Utility}(\mathcal{E}(\mathcal{G}, w)) \geq (\text{Utility}(\mathcal{G}) - \epsilon),
\end{aligned}
\end{equation}
where $\mathbb{P}[\cdot]$ denotes the probability of an event, $\mathcal{A}$ represents admissible attacks and $\mathcal{D}$ is the extraction function. $\text{Utility}(\cdot)$ quantifies graph quality, encompassing both structural fidelity and functional fidelity (e.g., performance on downstream tasks like link prediction). The parameter $\epsilon$ controls the utility preservation, ensuring that the watermarking process introduces only minimal degradation.

Instead of operating on the graph or its entangled representations, our solution decomposes the problem by learning orthogonal subspaces for structural and watermark information.

\subsubsection{Disentangled Encoder for Latent Representation Separation}
\label{sec:encoder}

Our method first addresses the utility-preservation constraint (Eq.~1). This is non-trivial for standard methods, whose use of a single latent space causes information entanglement. Such coupling means embedding a watermark $w$ inevitably perturbs the graph's utility-defining properties, directly conflicting with the constraint.

To resolve this, we turn to information theory. The data processing inequality provides a theoretical upper bound on this detrimental effect:
\begin{equation}
I(w; \text{Utility}) \leq I(h_w; h_s),
\end{equation}
where $h_s$ and $h_w$ structural representation and the watermark carrier, respectively. This inequality dictates that preserving utility requires minimizing the mutual information between the two representations. This forms the theoretical foundation for our encoder, which we formalize as the objective of learning maximally independent yet expressive representations:
\begin{equation}
\min I(h_s; h_w) \quad \text{subject to} \quad I(\mathcal{G}; h_s) \geq \gamma_s,\ I(\mathcal{G}; h_w) \geq \gamma_w,
\end{equation}
where $I(\cdot;\cdot)$ denotes the mutual information, and $\gamma_s, \gamma_w$ are thresholds that set a lower bound on the information that each representation must capture from the original graph $\mathcal{G}$.

To operationalize this principle, we design a disentangled encoder using a Graph Isomorphism Network (GIN) backbone, chosen for its expressive power and permutation invariance. The encoder uses separate projection heads to produce the structural representation $h_s$ and watermark carrier $h_w$. Architectural separation alone is insufficient, so we enforce our theoretical objective with a composite loss. First, to distill the graph's essential, utility-defining topological properties, we enforce invariance in $h_s$ via a contrastive objective:
\begin{equation}
\mathcal{L}_{\text{inv}} = \mathbb{E}_{\mathcal{G},\mathcal{G}'}[\|h_s(\mathcal{G}) - h_s(\mathcal{G}')\|_2^2],
\end{equation}
where $\mathcal{G}'$ indicate structurally perturbed variants of $\mathcal{G}$, generated through augmentations such as random node dropping and edge addition or deletion.
This trains the encoder to extract a stable, robust structural representation. Second, to minimize mutual information, we enforce orthogonality, a strong proxy for statistical independence:
\begin{equation}
\mathcal{L}_{\text{ortho}} = \|\mathbf{H}_s^\top \mathbf{H}_w\|^2,
\end{equation}
where $\mathbf{H}_s$ and $\mathbf{H}_w$ are batch-wise representation matrices. The complete objective, $\mathcal{L}_{\text{encoder}} = \mathcal{L}_{\text{inv}} + \lambda \mathcal{L}_{\text{ortho}}$, thus learns the desired representations. This theoretically-grounded approach ensures that modifications to the watermark carrier $h_w$ have a minimal effect on the invariant structural representation $h_s$, thereby satisfying our core constraint.

\subsubsection{Graph-aware INN for a High-Fidelity Channel}
\label{sec:inn}

Having addressed the utility-preservation constraint, we now focus on the primary objective from Eq.~1: maximizing watermark detection probability. The detectability of a watermark is fundamentally bounded by its signal integrity. Any information loss in the embedding function $\mathcal{E}$ would irreversibly degrade this signal, lowering the achievable detection rate even before an adversarial attack. Therefore, to maximize the theoretical bound on detectability, the embedding transformation must be perfectly reversible. This requirement directly compels the use of an INN, which is the deep learning operationalization of learnable, bijective functions.

However, a standard INN, while lossless, is not inherently robust. A generic, topology-agnostic transformation would embed a watermark that is brittle against graph-specific structural perturbations. To ensure that the watermark survives such attacks, the reversible transformation must be made \textbf{structurally aware}. We deduce that this can be achieved by conditioning the bijective mapping for the watermark carrier $h_w$ on the invariant structural representation $h_s$ learned by our encoder. This conditioning makes the watermark embedding sensitive to the graph's stable topological properties, which is the key to achieving robustness while reinforcing transparency by operating within the disentangled framework.

These deductions lead us to our proposed solution: a Graph-aware INN. We formalize it as a conditional INN that learns an invertible mapping $f_{\text{INN}}(h_w; h_s)$, which transforms the carrier distribution to a simple prior, $\mathcal{N}(0, I)$. The network is trained by minimizing the negative log-likelihood (NLL):
\begin{equation}
\mathcal{L}_{\text{NLL}} = \mathbb{E}_{h_w \sim \mathcal{G}} \left[ \frac{\|f_{\text{INN}}(h_w; h_s)\|_2^2}{2} - \log|\det J_{f_{\text{INN}}}(h_w; h_s)| \right],
\label{eq:inn_loss}
\end{equation}
where $J_{f_{\text{INN}}}$ is the transformation Jacobian. 
Architecturally, structural awareness is realized via a series of graph-aware coupling layers that compose the bijective mapping $f_{\text{INN}}$. 
This mapping is formed by a stack of alternating coupling layers where each layer equally splits its input $h_{w,in}$ into an unaltered part $h_{w,A}$ and a part $h_{w,B}$ that is affinely transformed using parameters from a GCN conditioned on $h_s$. The resulting concatenation, $h_{w,out}$, is passed to the next layer, with the stack's final output being the latent representation $z$.

The watermark injection is performed in the latent space. The carrier $h_w$ is first mapped to its latent representation, then shifted by the scaled watermark vector. This process is summarized as:
\begin{equation}
z_w = f_{\text{INN}}(h_w; h_s) + \alpha w,
\label{eq:watermark_injection}
\end{equation}
where $\alpha > 0$ is a hyperparameter controlling the embedding strength, which balances robustness and fidelity. This watermarked latent vector $z_w$ is then mapped back to the carrier space via the Graph-aware INN's inverse to produce the modified carrier, $\tilde{h}_w = f_{\text{INN}}^{-1}(z_w; h_s)$, for the structure-aware editor.

\subsubsection{Structure-aware Editor for Controllable Discretization}
\label{sec:editor}

A key challenge is translating the continuous watermarked carrier $\tilde{h}_w$ from preceding modules into a discrete graph $\mathcal{G}_w$. A standard graph decoder is unsuitable for this task. Such decoders are trained for reconstruction fidelity, causing them to treat the low-magnitude watermark signal as noise to be discarded during the lossy quantization to a discrete adjacency matrix. This failure mode, termed discretization-induced watermark degradation, would catastrophically compromise watermark robustness.

The failure of standard decoders necessitates a purpose-built module with an entirely different objective: not graph reconstruction, but maximal watermark preservation. To achieve this robustly, we deduce its decisions must be guided by two information sources: the modified carrier  to determine what changes to embed, and the invariant structural representation $h_s$ to identify the most stable topological locations of where to embed them. This leads to the formulation of an optimal editing problem, which seeks an edit plan $\Delta E^*$ that maximizes expected watermark detectability under attack, subject to a budget constraint $k$:
\begin{equation}
\Delta E^* = \arg \max_{\Delta E, |\Delta E| \le k} \mathbb{E}_{\mathcal{A}}[\mathcal{T}(\mathcal{G} \oplus \Delta E, w)],
\label{eq:editor_objective}
\end{equation}
where $\mathcal{T}$ is the detection statistic and $\oplus$ denotes the symmetric difference, which applies the edge edits in $\Delta E$ to the graph $\mathcal{G}$. As this objective is intractable, we implement a structure-aware editor, $\Phi_D$, as a neural network trained to approximate this function. Concretely, the editor's dyadic affinity predictor computes an editing score $s_{uv}=\text{MLP}([h_s(u) \| \tilde{h}_w(u) \| h_s(v) \| \tilde{h}_w(v)])$ for a scalable set of candidate edges (all existing edges and a random sample of non-edges).These scores are then ranked to generate a binary Top-k Mask matrix, where only the entries for the top $k$ scores are set to 1. This mask is subsequently applied to the original adjacency matrix to perform edge flips, and the set of these $k$ modifications constitutes the discrete edit plan $\Delta E$.

\subsection{Watermark Verification}
\label{sec:verification}

Verification must statistically determine if a suspect graph contains the owner's watermark $w$. Given the risk of adversarial attacks and the gravity of false accusations, an ad-hoc rule is insufficient. Trustworthy attribution with quantifiable confidence therefore necessitates formulating verification as a statistical hypothesis test.

\subsubsection{Statistical Hypothesis Framework}

A hypothesis test is defined by a null ($H_0$, watermark absent) and alternative ($H_1$, watermark present) hypothesis. Crucially, these are not arbitrary assumptions but are emergent properties of our framework's design. The Graph-aware INN is trained to map any unwatermarked graph's carrier to a standard Gaussian prior, which directly defines the null hypothesis for an extracted latent representation $z_{\text{test}}$:
\begin{equation}
H_0: z_{\text{test}} \sim \mathcal{N}(0, I)
\end{equation}
Conversely, our embedding process injects the watermark by systematically shifting the distribution's mean. The alternative hypothesis is therefore defined by the presence of this shift, potentially attenuated by attacks:
\begin{equation}
H_1: z_{\text{test}} \sim \mathcal{N}(\alpha' w, I),
\end{equation}
where $\alpha' \leq \alpha$ accounts for potential signal degradation.

\subsubsection{Optimal Detection Statistic}

With two simple hypotheses defined, statistical theory dictates the optimal procedure for distinguishing between them. The Neyman-Pearson lemma states that the most powerful test is the likelihood ratio test (LRT), compelling its use to achieve maximum detection power. For our specific case of two Gaussian distributions differing only by their mean, the log-LRT simplifies algebraically to a matched filter. This yields our optimal detection statistic, 
$\hat{\mathcal{T}} = \langle z_{\text{test}}, w \rangle$, calculated as the inner product between the extracted signal $z_{\text{test}}$ and the watermark $w$.
This statistic maximizes the signal-to-noise ratio by projecting the extracted signal onto the known watermark direction. The distributions of $\hat{\mathcal{T}}$ under each hypothesis are consequently well-characterized:
\begin{align}
\hat{\mathcal{T}} \mid H_0 &\sim \mathcal{N}(0, \|w\|^2) \\
\hat{\mathcal{T}} \mid H_1 &\sim \mathcal{N}(\alpha' \|w\|^2, \|w\|^2)
\end{align}

\begin{table*}[t!]
\centering
\caption{Main results on watermark detectability and robustness (AUC). Results are averaged by category, with the best performance in \textbf{bold}. The '-' indicates that latent-space adversarial attacks were not applicable to the structure-space methodology of Towards~\cite{zhao2015towards}.}

\label{tab:main_results}
\vspace{-1em}
\small 
\setlength{\tabcolsep}{12.9pt}
\resizebox{\textwidth}{!}{%
\begin{tabular}{l|l|c|ccc|ccc|cc|c}
\toprule
\multirow{2}{*}{\textbf{Category}} & \multirow{2}{*}{\textbf{Method}} & \multirow{2}{*}{\textbf{Clean}} & \multicolumn{3}{c|}{\textbf{Edge Flip.}} & \multicolumn{3}{c|}{\textbf{Node Del.}} & \multicolumn{2}{c|}{\textbf{Adversarial Atk.}} & \multirow{2}{*}{\textbf{Iso. Var.}} \\
\cmidrule{4-11}
& & & \textbf{10\%} & \textbf{30\%} & \textbf{50\%} & \textbf{10\%} & \textbf{30\%} & \textbf{50\%} & \textbf{NEA} & \textbf{L2 Metric} & \\
\midrule
\multirow{4}{*}{Social Networks}
& Towards~\cite{zhao2015towards} & 0.981 & 0.753 & 0.612 & 0.521 & 0.715 & 0.584 & 0.509 & - & - & 0.821 \\
& KGMark~\cite{peng2025kgmark} & 0.998 & 0.990 & 0.973 & 0.938 & 0.981 & 0.952 & 0.915 & 0.948 & 0.942 & 0.994 \\
& Naive Baseline & 0.997 & 0.942 & 0.768 & 0.598 & 0.908 & 0.709 & 0.572 & 0.738 & 0.731 & 0.991 \\
& \textbf{DRGW} & \textbf{0.999} & \textbf{0.994} & \textbf{0.984} & \textbf{0.962} & \textbf{0.989} & \textbf{0.975} & \textbf{0.956} & \textbf{0.958} & \textbf{0.955} & \textbf{0.998} \\
\midrule
\multirow{4}{*}{Web Graphs}
& Towards~\cite{zhao2015towards} & 0.975 & 0.721 & 0.585 & 0.510 & 0.695 & 0.561 & 0.505 & - & - & 0.805 \\
& KGMark~\cite{peng2025kgmark} & 0.997 & 0.988 & 0.969 & 0.932 & 0.978 & 0.948 & 0.909 & 0.943 & 0.938 & 0.993 \\
& Naive Baseline & 0.996 & 0.935 & 0.752 & 0.584 & 0.899 & 0.691 & 0.553 & 0.721 & 0.714 & 0.990 \\
& \textbf{DRGW} & \textbf{0.998} & \textbf{0.992} & \textbf{0.981} & \textbf{0.959} & \textbf{0.986} & \textbf{0.971} & \textbf{0.953} & \textbf{0.952} & \textbf{0.949} & \textbf{0.997} \\
\midrule
\multirow{4}{*}{Academic Networks}
& Towards~\cite{zhao2015towards} & 0.983 & 0.766 & 0.634 & 0.531 & 0.725 & 0.597 & 0.513 & - & - & 0.835 \\
& KGMark~\cite{peng2025kgmark} & 0.999 & 0.992 & 0.977 & 0.945 & 0.984 & 0.959 & 0.924 & 0.957 & 0.953 & 0.996 \\
& Naive Baseline & 0.998 & 0.955 & 0.789 & 0.629 & 0.926 & 0.741 & 0.598 & 0.769 & 0.761 & 0.994 \\
& \textbf{DRGW} & \textbf{0.999} & \textbf{0.995} & \textbf{0.986} & \textbf{0.970} & \textbf{0.992} & \textbf{0.981} & \textbf{0.965} & \textbf{0.966} & \textbf{0.964} & \textbf{0.999} \\
\midrule
\multirow{4}{*}{Knowledge Graphs}
& Towards~\cite{zhao2015towards} & 0.978 & 0.740 & 0.601 & 0.515 & 0.703 & 0.578 & 0.508 & - & - & 0.817 \\
& KGMark~\cite{peng2025kgmark} & 0.998 & 0.989 & 0.971 & 0.935 & 0.979 & 0.950 & 0.912 & 0.947 & 0.941 & 0.993 \\
& Naive Baseline & 0.997 & 0.939 & 0.764 & 0.593 & 0.905 & 0.703 & 0.565 & 0.732 & 0.725 & 0.990 \\
& \textbf{DRGW} & \textbf{0.999} & \textbf{0.993} & \textbf{0.982} & \textbf{0.964} & \textbf{0.987} & \textbf{0.974} & \textbf{0.957} & \textbf{0.957} & \textbf{0.954} & \textbf{0.998} \\
\midrule
\multirow{4}{*}{E-commerce \& Rec.}
& Towards~\cite{zhao2015towards} & 0.980 & 0.749 & 0.609 & 0.519 & 0.708 & 0.581 & 0.509 & - & - & 0.819 \\
& KGMark~\cite{peng2025kgmark} & 0.998 & 0.989 & 0.972 & 0.936 & 0.980 & 0.951 & 0.913 & 0.949 & 0.944 & 0.994 \\
& Naive Baseline & 0.997 & 0.943 & 0.770 & 0.600 & 0.912 & 0.712 & 0.570 & 0.739 & 0.732 & 0.991 \\
& \textbf{DRGW} & \textbf{0.999} & \textbf{0.994} & \textbf{0.983} & \textbf{0.958} & \textbf{0.988} & \textbf{0.976} & \textbf{0.951} & \textbf{0.959} & \textbf{0.956} & \textbf{0.998} \\
\midrule
\multirow{4}{*}{Road Networks}
& Towards~\cite{zhao2015towards} & 0.520 & 0.522 & 0.514 & 0.526 & 0.518 & 0.511 & 0.524 & - & - & 0.523 \\
& KGMark~\cite{peng2025kgmark} & 0.996 & 0.984 & 0.961 & 0.924 & 0.971 & 0.939 & 0.899 & 0.935 & 0.930 & 0.990 \\
& Naive Baseline & 0.994 & 0.921 & 0.725 & 0.555 & 0.880 & 0.668 & 0.528 & 0.699 & 0.692 & 0.986 \\
& \textbf{DRGW} & \textbf{0.998} & \textbf{0.985} & \textbf{0.964} & \textbf{0.932} & \textbf{0.978} & \textbf{0.945} & \textbf{0.908} & \textbf{0.941} & \textbf{0.937} & \textbf{0.996} \\
\bottomrule
\end{tabular}%
}
\end{table*}

\subsubsection{Decision Rule with Controlled Error Rates}

The final step is to set a decision threshold for $\hat{\mathcal{T}}$. In ownership verification, the most critical error to control is a false positive. We therefore adopt the Neyman-Pearson testing framework, which allows us to fix the maximum acceptable false positive rate $\beta$. Based on the known distribution of $\hat{\mathcal{T}}$ under $H_0$, this pre-defined error rate 
allows us to compute a precise decision threshold $\tau = \|w\| \cdot \Phi^{-1}(1 - \beta)$,
where $\Phi^{-1}$ is the standard normal quantile function. For any observed statistic $\tau_{\text{obs}}$, this is equivalent to computing a $p$-value:
\begin{equation}
p = 1 - \Phi\left(\frac{\tau_{\text{obs}}}{\|w\|}\right)
\end{equation}
The final decision rule is to reject $H_0$ (confirming watermark presence) if and only if $p < \beta$. This statistically-grounded approach provides strong protection against erroneous attribution while maximizing detection power.

\section{Experimental Setup}

\textbf{Datasets.} We conduct a comprehensive evaluation on 18 real-world graphs spanning 6 diverse categories: Social Networks, Academic Networks, Knowledge Graphs, E-commerce \& Recommendation, Web Graphs, and Road Networks. This selection allows for a thorough assessment of DRGW's performance across graphs with distinct structural properties, from the scale-free nature of social networks to the grid-like topology of road networks. Detailed statistics for each dataset are provided in Appendix~\ref{app:datasets}. To ensure a clear presentation, all experimental results reported in the main paper are the average metrics for each category.

\textbf{Baselines.} We compare DRGW against the state-of-the-art methods from the two dominant paradigms. The first is \textbf{Towards}~\cite{zhao2015towards}, a seminal work representing the structural-space paradigm. The second is \textbf{KGMark}~\cite{peng2025kgmark}, the current state-of-the-art from the entangled latent-space paradigm. To empirically validate our central hypotheses, we also design a critical variant, \textbf{Naive Latent-Space Baseline}, which replaces our specialized components with a standard GNN autoencoder and decoder. For a fair comparison, we utilized the official implementations for all baselines where available and adopted the hyperparameter settings reported in their original papers. 

\textbf{Evaluation Metrics.} We evaluate our framework on three core properties: \textbf{Detectability}, \textbf{Robustness}, and \textbf{Transparency}. Detectability and robustness are measured using the Area Under the ROC Curve (AUC). We assess transparency via two perspectives: \textit{Structural Fidelity}, using Flipped Edges (\%), Assortativity, Global Clustering Coefficient, and dK-2 Deviation~\cite{zhao2015towards}; and \textit{Functional Fidelity}, using Performance Drop (\%) on link prediction and the Cosine Similarity of node embeddings. For all Transparency metrics, lower values indicate better performance, except for Cosine Similarity. Detailed explanations of these metrics are provided in Appendix~\ref{app:metrics}.

\begin{table*}[t!]
\centering
\caption{Framework transparency. Lower values are better (indicated by ↓) except for Cosine Similarity (↑). The '-' indicates the Towards~\cite{zhao2015towards} method was inapplicable to the regular structure of Road Networks.}

\vspace{-1em}
\label{tab:transparency}
\small 
\setlength{\tabcolsep}{10.5pt}
\resizebox{\textwidth}{!}{
\begin{tabular}{l|l|cccc|cc}
\toprule
\multirow{2}{*}{\textbf{Category}} & \multirow{2}{*}{\textbf{Method}} & \multicolumn{4}{c|}{\textbf{Structural Fidelity}} & \multicolumn{2}{c}{\textbf{Functional Fidelity}} \\
\cmidrule{3-8}
& & \textbf{Flipped Edges (\%)} & \textbf{Assortativity (\%)} & \textbf{Clustering Coefficient(\%)} & \textbf{dK-2 Dev.} & \textbf{Link Pred. (\%)} & \textbf{Cosine Sim.} \\
\midrule
\multirow{4}{*}{Social Networks}
& Towards~\cite{zhao2015towards} & \textbf{0.002} & 5.12 & 4.35 & 0.085 & 4.13 & 0.721 \\
& KGMark~\cite{peng2025kgmark} & 0.087 & 2.89 & 2.11 & 0.041 & 2.35 & 0.970 \\
& Naive Baseline & 0.101 & 3.15 & 2.45 & 0.048 & 2.85 & 0.965 \\
& \textbf{DRGW} & 0.058 & \textbf{0.48} & \textbf{0.41} & \textbf{0.015} & \textbf{0.72} & \textbf{0.993} \\
\midrule
\multirow{4}{*}{Web Graphs}
& Towards~\cite{zhao2015towards} & \textbf{0.003} & 6.03 & 5.18 & 0.102 & 4.81 & 0.685 \\
& KGMark~\cite{peng2025kgmark} & 0.083 & 3.52 & 3.01 & 0.055 & 2.95 & 0.955 \\
& Naive Baseline & 0.099 & 3.95 & 3.45 & 0.062 & 3.45 & 0.948 \\
& \textbf{DRGW} & 0.062 & \textbf{0.65} & \textbf{0.58} & \textbf{0.021} & \textbf{1.04} & \textbf{0.989} \\
\midrule
\multirow{4}{*}{Academic Networks}
& Towards~\cite{zhao2015towards} & \textbf{0.002} & 4.88 & 4.10 & 0.079 & 3.88 & 0.735 \\
& KGMark~\cite{peng2025kgmark} & 0.086 & 2.65 & 2.01 & 0.039 & 2.15 & 0.972 \\
& Naive Baseline & 0.091 & 2.95 & 2.35 & 0.045 & 2.65 & 0.968 \\
& \textbf{DRGW} & 0.059 & \textbf{0.39} & \textbf{0.35} & \textbf{0.013} & \textbf{0.68} & \textbf{0.994} \\
\midrule
\multirow{4}{*}{Knowledge Graphs}
& Towards~\cite{zhao2015towards} & \textbf{0.002} & 5.05 & 4.21 & 0.081 & 3.95 & 0.710 \\
& KGMark~\cite{peng2025kgmark} & 0.088 & 2.81 & 2.08 & 0.040 & 2.25 & 0.971 \\
& Naive Baseline & 0.093 & 3.15 & 2.45 & 0.047 & 2.75 & 0.967 \\
& \textbf{DRGW} & 0.057 & \textbf{0.52} & \textbf{0.43} & \textbf{0.014} & \textbf{0.70} & \textbf{0.994} \\
\midrule
\multirow{4}{*}{E-commerce \& Rec.}
& Towards~\cite{zhao2015towards} & \textbf{0.003} & 5.33 & 4.52 & 0.089 & 4.28 & 0.702 \\
& KGMark~\cite{peng2025kgmark} & 0.084 & 3.01 & 2.24 & 0.044 & 2.50 & 0.968 \\
& Naive Baseline & 0.099 & 3.35 & 2.65 & 0.050 & 3.00 & 0.964 \\
& \textbf{DRGW} & 0.061 & \textbf{0.55} & \textbf{0.49} & \textbf{0.017} & \textbf{0.81} & \textbf{0.992} \\
\midrule
\multirow{4}{*}{Road Networks}
& Towards~\cite{zhao2015towards} & - & - & - & - & - & - \\
& KGMark~\cite{peng2025kgmark} & 0.080 & 5.82 & 5.30 & 0.095 & 5.30 & 0.928 \\
& Naive Baseline & 0.097 & 6.25 & 5.79 & 0.101 & 5.80 & 0.922 \\
& \textbf{DRGW} & 0.065 & \textbf{1.85} & \textbf{1.71} & \textbf{0.058} & \textbf{2.45} & \textbf{0.975} \\
\bottomrule
\end{tabular}%
}
\end{table*}

\textbf{Implementation Details.} Our framework is implemented using PyTorch and PyG. To ensure reproducibility, a complete and detailed description of the model architectures, training protocol, and all hyperparameter settings is provided in Appendix~\ref{app:implementation}.

\subsection{Detectability}
We begin by evaluating the watermark detectability of our watermark on clean, unattacked graphs, with the results summarized in the clean column of Table~\ref{tab:main_results}. In the absence of attacks, DRGW achieves near-perfect detectability. Across all six dataset categories, DRGW consistently achieves an AUC of 0.998 or higher. This result is a direct validation of our framework: Graph-aware INN provides a lossless channel that maximizes the watermark's signal integrity in the latent space, while the structure-aware editor ensures this high-fidelity signal is successfully translated into a detectable pattern in the final discrete graph.

 Our evaluation reveals critical insights into the characteristics of different watermarking schemes. \textbf{KGMark}~\cite{peng2025kgmark} also shows excellent clean performance, underscoring the potential of latent-space methods under ideal conditions. More revealingly, our \textbf{Naive Baseline} achieves a deceptively high AUC. This highlights a crucial limitation of evaluating detectability in isolation: a simple autoencoder can create a detectable—albeit fragile—watermark when no adversarial pressure is applied. 
In stark contrast, the structural-space method, \textbf{Towards}~\cite{zhao2015towards}, exhibits a significant vulnerability. While performing adequately on graphs with complex local topologies like Social Networks (AUC 0.981), its performance plummets to random guessing on Road Networks (AUC 0.520). 
This failure shows a fundamental limitation of the method itself: its reliance on specific subgraph patterns makes it inapplicable to graphs with regular, grid-like structures. 

\subsection{Transparency}
\label{sec:transparency}
A central failure of existing latent-space paradigms is the entanglement between a graph's functional properties and the watermark signal. This section quantitatively evaluates whether if the DRGW's principled design resolves this challenge. The results in Table~\ref{tab:transparency} empirically confirm the success of this approach. DRGW's disentanglement-by-design achieves state-of-the-art transparency. Across all metrics, DRGW reduces fidelity loss by an order of magnitude compared to entangled methods. The most striking contrast is with the Naive Baseline, which directly exposes the cost of entanglement. On Social Networks, DRGW slashes the functional performance drop on link prediction to a mere 0.72\%, a nearly 75\% reduction compared to the 2.85\% drop from the Naive Baseline. This is the direct, measurable outcome of our architecture: by confining all watermark-related perturbations to an orthogonal subspace, the foundational structural and functional information encoded in $h_s$ is effectively shielded from corruption.

A crucial implication of this principled disentanglement is not just superior performance, but also \textbf{predictable controllability} over the trade-off between transparency and robustness. To investigate this, we conducted an extensive sensitivity analysis by varying the watermark strength ($\alpha$) and the editing budget ($k$). We found that the fidelity loss increases smoothly and monotonically with both parameters, forming a predictable cost surface rather than a chaotic one. Conversely, robustness exhibits a pattern of diminishing returns, quickly reaching a broad plateau of near-optimal performance. The key insight from this analysis is the existence of a wide "sweet spot" where high robustness can be achieved at a minimal cost to fidelity. Our chosen default parameters ($\alpha=0.1, k=0.1\%$) fall squarely within this optimal region, validating that our architecture elegantly resolves the trade-off between these two critical objectives. The detailed heatmaps visualizing this analysis are provided in Appendix~\ref{app:sensitivity}.

\subsection{Robustness}
\label{sec:robustness} 
To evaluate the framework's robustness against the adaptive adversary defined in our threat model, we use structural attacks (edge/node modifications, isomorphism variations) and two sophisticated adaptive attack: \textbf{NEA}~\cite{bojchevski2019adversarial}, a node embedding poisoning attack, and the \textbf{L2 Metric} attack~\cite{bhardwaj2021adversarial}, which perturbs embeddings via instance attribution. As shown in Table~\ref{tab:main_results}, DRGW is highly resilient. For instance, under a 50\% node deletion on Academic Networks, it maintains a 0.970 AUC, consistently outperforming all baselines.

The necessity of our framework is evidenced by the catastrophic failure of the \textbf{Naive Baseline}. While deceptively effective on clean data, its performance collapses under pressure, with its AUC plummeting to just 0.598 under a 50\% edge flip on Social Networks—barely better than random guessing. Its standard decoder, optimized for reconstruction, cannot distinguish the faint watermark signal from attack-induced noise, effectively erasing the watermark during the lossy quantization back to a discrete graph.

In stark contrast, DRGW's robustness is a collaborative achievement of a two-part mechanism designed to overcome this exact failure mode. First, the Disentangled Encoder provides an invariant structural representation, which acts as a stable "anchor" or coordinate system that remains largely consistent even as the graph is perturbed. Second, and crucially, the structure-aware Editor leverages this stable anchor. It is purpose-built not for reconstruction, but to intelligently translate the latent watermark into a set of durable, targeted discrete edits. By using the invariant structural information to decide \textit{where} to place modifications for maximum resilience, it ensures the watermark's physical footprint survives both the discretization process and subsequent attacks.

To visually corroborate this resilience, we provide a qualitative analysis of the carrier subspace ($h_w$) under attack in Appendix~\ref{app:robustness_vis}. As shown in the visualizations therein, even after severe corruption via 50\% node deletion and 50\% edge flipping, the watermarked distribution retains its macroscopic stream-like structure. While the clusters become more diffuse, their integrity is preserved, providing compelling visual evidence that our editor embeds a resilient, physical watermark.

\subsection{Ablation Study}
To provide a causal analysis of DRGW's success, we conduct an ablation study by deconstructing its key components. The results in Table~\ref{tab:ablation} validate that DRGW's performance stems from its design, not incidental complexity.

We first test the necessity of the \textbf{Disentangled Encoder} by removing its mutual information minimization objective in the \textit{w/o Disentanglement} variant. This re-introduces information entanglement, and the consequences are immediate: the transparency loss (Trans.) skyrockets from 0.72\% to 2.85\%, a nearly four-fold increase. This provides strong causal evidence that our encoder's ability to separate functional and watermark representations is the foundational reason for DRGW's high transparency. This mechanism is visually corroborated by the qualitative analysis in Appendix~\ref{app:disentanglement_vis}, which shows a clear separation of our structural ($h_s$) and carrier ($h_w$) subspaces in contrast to the mixed representations from the entangled baseline. This finding is further validated quantitatively in Appendix~\ref{app:disentanglement_quant}.

Next, we validate the role of the \textbf{structure-aware Editor} by replacing it with a standard graph decoder in the \textit{w/o Editor} variant. This simulates the "uncontrollable discretization" failure mode. The model's robustness (Rob.) under attack catastrophically degrades, with the AUC plummeting from 0.984 to 0.791. This proves our purpose-built editor is critical for translating latent signals into durable graph edits, ensuring final robustness and detectability.

Finally, we assess the contribution of the \textbf{Graph-aware INN}. Removing this module in the \textit{w/o INN} variant also leads to a severe drop in robustness (AUC to 0.895). This confirms that the INN's lossless channel is critical for maximizing the watermark's signal-to-noise ratio. A weaker, noisier signal compromises the editor's ability to perform robust edits, thus degrading the entire system's performance on both detectability and robustness.
\begin{table}[htbp]
\centering
\small 
\caption{Causal analysis of DRGW's components. We report robustness (Rob.) as the average AUC under the 30\% Edge Flip attack, and transparency (Trans.) as the average functional fidelity loss (\%). Arrows indicate that a lower (↓) or higher (↑) value is worse.}
\label{tab:ablation}
\vspace{-1em}
\begin{tabular}{l c c}
\toprule
\textbf{Variant} & \textbf{Rob. (AUC) ↓} & \textbf{Trans. (Loss \%) ↑} \\ 
\midrule
\textbf{DRGW (Full Model)} & \textbf{0.984} & \textbf{0.72} \\
\midrule
w/o Disentanglement & 0.976 & 2.85 \\
w/o Graph-aware INN & 0.895 & 0.95 \\
w/o Editor & 0.791 & 3.10 \\
\bottomrule
\end{tabular}
\vspace{-0.5cm}
\end{table}

\section{Discussion}
\label{sec:discussion}

DRGW's topology-agnostic nature grants it superior generality over structure-space methods, enabling application to diverse graphs like road networks. However, this generality comes at the cost of higher computational complexity than simplistic benchmarks (e.g., Towards~[39]), though it remains significantly more efficient than generative latent-space competitors (e.g., KGMark~[26]), as its single forward-pass design avoids the costly, multi-step iterative sampling required by diffusion-based decoders. This positions DRGW favorably in the trade-off space between robustness and efficiency.

The effective watermark capacity is bounded not by the watermark dimension $w$ alone, but by the expressivity of the carrier subspace $h_w$ and the editor's budget $k$. Our framework's design focuses on graph topology, demonstrating robustness even on heterogeneous data like knowledge graphs. While effective, a compelling direction for future work is to explicitly model rich node and edge semantics. This requires developing type-aware disentanglement to isolate the watermark from complex, type-specific information, which would further enhance transparency and robustness in heterogeneous environments.

\section{Conclusion}
In this paper, we introduce DRGW, a novel watermarking framework for graph-structured data, leveraging disentangled representation learning to ensure robustness, transparency, and detectability. Our method addresses the fundamental challenges of information entanglement and uncontrollable discretization in graph watermarking, offering a secure solution that maintains structural and functional fidelity even under adversarial environments. DRGW provides a foundation for securing the integrity and ownership of graph data assets, with potential applications spanning academic research to commercial deployments in fields like social network analysis and recommendation systems.

\begin{acks}
This work was  supported by the Strategic Priority Research Program of the Chinese Academy of Sciences (NO.XDB0690302),  NSFC under grant No.62371450 and Project of Key Laboratory Cyberspace Security Defense under grant No.2025-A03.

\end{acks}
\clearpage
\bibliographystyle{ACM-Reference-Format}
\balance
\bibliography{reference}

\clearpage 

\appendix

\section{Dataset Details}
\label{app:datasets}
This section provides detailed statistics for the 18 datasets used in our experiments, as summarized in Table~\ref{tab:datasets}. Our selection covers 6 major categories to ensure a comprehensive evaluation. These datasets vary significantly in scale, from thousands to millions of nodes and edges, and encompass a wide range of topological characteristics, providing a robust testbed for our watermarking framework.

\begin{table}[htbp]
\centering
\caption{Overview of the 18 datasets across 6 categories used for evaluation.}
\label{tab:datasets}
\small{
\begin{tabular}{llrr}
\toprule
\textbf{Category} & \textbf{Dataset} & \textbf{Nodes} & \textbf{Edges} \\
\midrule
\multirow{4}{*}{Social Networks} & Facebook (ego-net) & 4,039 & 88,234 \\
& Epinions & 75,879 & 405,740 \\
& Pokec & 1,632,803 & 22,301,964 \\
& LiveJournal & 4,847,571 & 68,993,773 \\
\midrule
\multirow{3}{*}{Academic Networks} & Patents & 23,133 & 93,468 \\
& ogbn-arxiv & 169,343 & 1,166,243 \\
& DBLP & 317,080 & 1,049,866 \\
\midrule
\multirow{3}{*}{Knowledge Graphs} & MIND & 24,733 & 148,568 \\
& YAGO3-10 & 123,182 & 1,089,040 \\
& ogbl-wikikg2 & 2,500,604 & 17,137,170 \\
\midrule
\multirow{3}{*}{E-commerce \& Rec.} & Last-FM & 58,266 & 464,567 \\
& Amazon Co-purchasing & 262,111 & 899,792 \\
& ogbn-products & 2,449,029 & 61,238,448 \\
\midrule
\multirow{3}{*}{Web Graphs} & web-Stanford & 281,903 & 1,992,636 \\
& BerkStan & 685,230 & 6,649,470 \\
& web-Google & 875,713 & 4,322,051 \\
\midrule
\multirow{2}{*}{Road Networks} & roadNet-TX & 1,379,917 & 1,921,660 \\
& roadNet-CA & 1,965,206 & 2,766,607 \\
\bottomrule
\end{tabular}%
}
\end{table}

\section{Implementation Details and Training Protocol}
\label{app:implementation}
\subsection{Model Architecture}
The encoder $\Phi_E$ employs a 4-layer Graph Isomorphism Network (GIN) with 256-dimensional hidden states, followed by separate linear projection heads to produce the structural ($h_s$) and watermark ($h_w$) subspaces. The Graph-aware INN $f_{\text{INN}}$ consists of 8 affine coupling layers. The scaling and translation parameters within these layers are computed by a conditioning network, which is a 2-layer graph neural network that takes $h_s$ as input. The editor $\Phi_D$ is a 3-layer MLP with GELU activations that computes pairwise editing scores. All latent representations ($h_s$, $h_w$) are 256-dimensional.

\subsection{Training Protocol and Hyperparameters}
The framework is trained in an end-to-end way using the AdamW optimizer with $(\beta_1, \beta_2) = (0.9, 0.999)$ and a weight decay of $1\times10^{-4}$. The learning rate follows a cosine decay schedule from an initial value of $1\times10^{-3}$ down to $1\times10^{-5}$. The training follows a three-stage curriculum with a total of 6000 epochs, using a batch size of 32.

\textbf{Stage 1: Disentangled Representation Learning (1000 epochs).}
The encoder $\Phi_E$ is trained to learn an invariant structural representation $h_s$ and an independent watermark carrier $h_w$. The objective is:
\begin{equation}
\mathcal{L}_{\text{stage1}} = \mathcal{L}_{\text{inv}} + \lambda_{\text{MI}}\mathcal{L}_{\text{ortho}},
\end{equation}
where $\mathcal{L}_{\text{inv}}$ is the contrastive loss on augmented graphs and $\mathcal{L}_{\text{ortho}}$ is the orthogonality constraint. The loss weight is set to $\lambda_{\text{MI}} = 0.1$.

\textbf{Stage 2: System Initialization via Reconstruction (2000 epochs).}
With the encoder's weights frozen, the INN $f_{\text{INN}}$ and editor $\Phi_D$ are pre-trained on a graph reconstruction task. The objective is:
\begin{equation}
\mathcal{L}_{\text{stage2}} = \lambda_{\text{recon}}\mathcal{L}_{\text{recon}} + \lambda_{\text{inn}}\mathcal{L}_{\text{NLL}} + \lambda_{\text{cycle}}\mathcal{L}_{\text{cycle}},
\end{equation}
where $\mathcal{L}_{\text{recon}}$ is the binary cross-entropy reconstruction loss, and $\mathcal{L}_{\text{cycle}}$ is a cycle-consistency loss for the INN. The weights are set to $\lambda_{\text{recon}} = 1.0$, $\lambda_{\text{inn}} = 5.0$, and $\lambda_{\text{cycle}} = 10.0$.

\textbf{Stage 3: Adversarial Robustness Fine-tuning (3000 epochs).}
Finally, the entire system is fine-tuned to maximize watermark survivability against attacks. The training objective shifts to:
\begin{equation}
\mathcal{L}_{\text{stage3}} = \lambda_{\text{robust}}\mathcal{L}_{\text{robust}} + \lambda_{\text{inn}}\mathcal{L}_{\text{NLL}}
\end{equation}
The adversarial loss, $\mathcal{L}_{\text{robust}} = -\mathbb{E}[\mathcal{T}(\mathcal{G}'_w, w)]$, directly maximizes the expected watermark detection statistic $\mathcal{T}$ on attacked graphs. The weights for this stage are $\lambda_{\text{robust}} = 1.0$, while the INN loss weight $\lambda_{\text{inn}}$ remains $5.0$.

Key watermark-specific parameters are fixed across all experiments: the watermark dimension is 128, the embedding strength is $\alpha=0.1$, and the graph editing budget is $k = 0.1\%$ of the total number of edges.

\section{Parameter Sensitivity Analysis}
\label{app:sensitivity}

\begin{figure}[htbp]
    \centering
    \begin{subfigure}[b]{0.49\columnwidth}
        \centering
        \includegraphics[width=\textwidth]{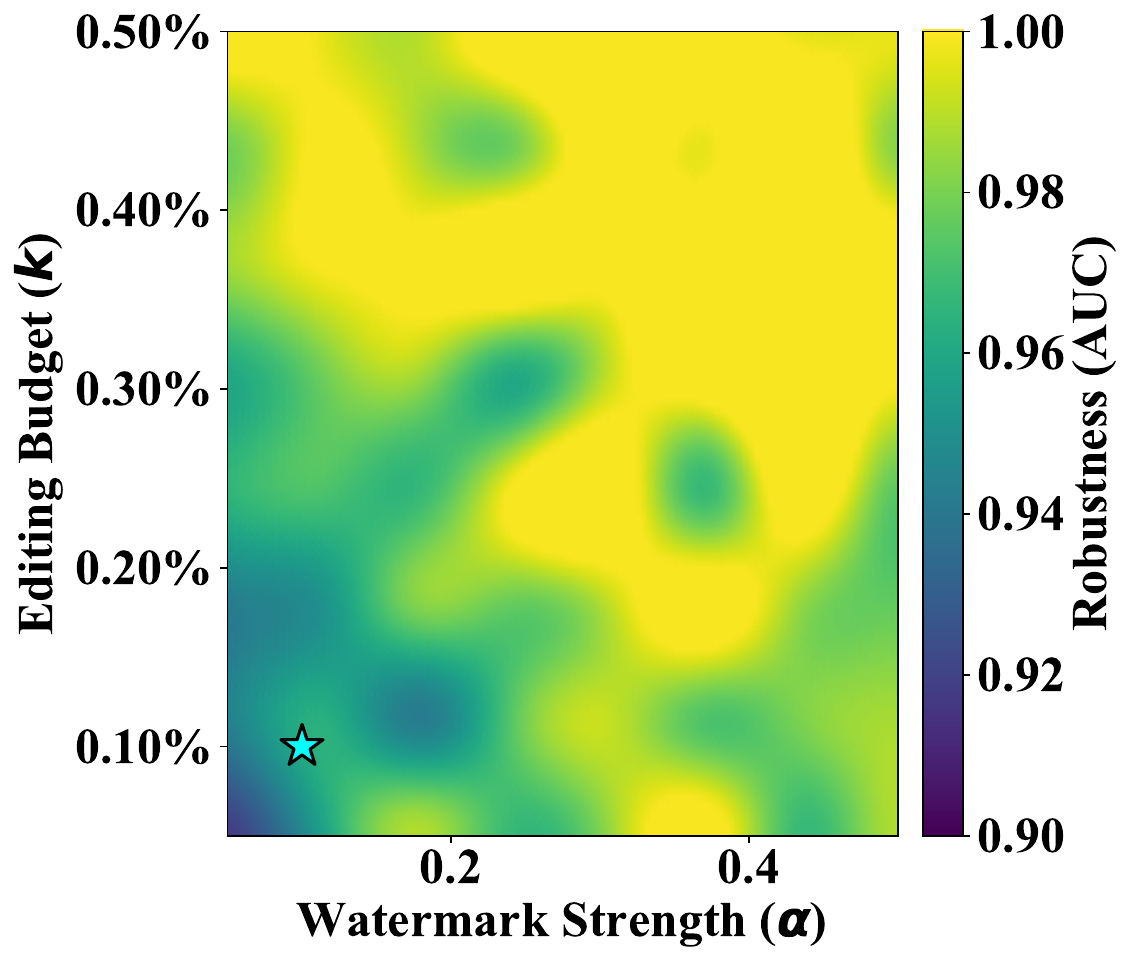}
        \caption{Robustness (AUC)}
        \label{fig:sensitivity_robustness_appendix}
    \end{subfigure}
    \hfill
    \begin{subfigure}[b]{0.49\columnwidth}
        \centering
        \includegraphics[width=\textwidth]{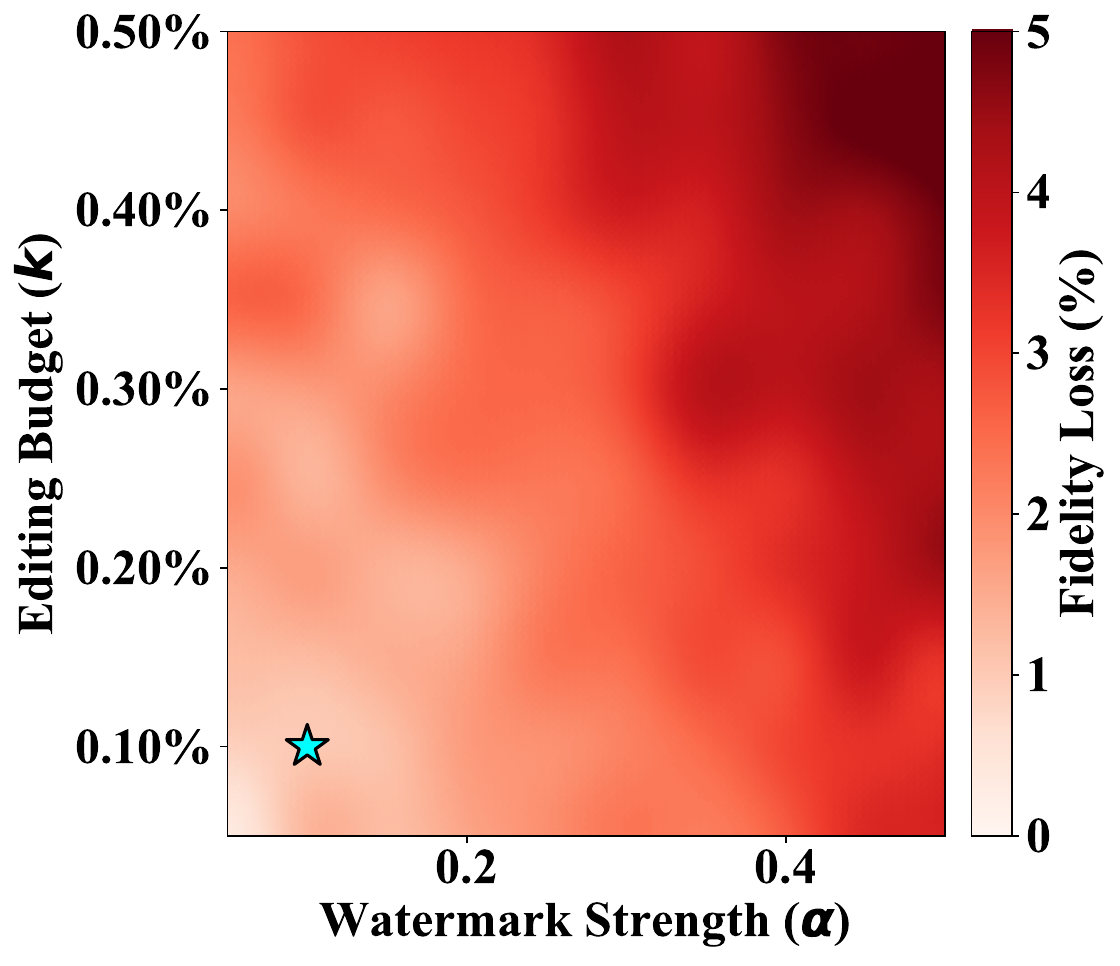}
        \caption{Fidelity Loss (\%)}
        \label{fig:sensitivity_fidelity_appendix}
    \end{subfigure}
    \caption{Analysis of Controllability. The heatmaps illustrate the trade-off between (a) Robustness and (b) Fidelity Loss as a function of watermark strength ($\alpha$) and editing budget ($k$). The star indicates our default operating point, which offers high robustness for low fidelity loss.}
    \label{fig:param_sensitivity_appendix}
\end{figure}

To provide a deeper understanding of the interplay between watermark robustness and transparency, we present a detailed parameter sensitivity analysis in Figure~\ref{fig:param_sensitivity_appendix}. The heatmaps visualize the trade-off between Robustness (measured in AUC) and Fidelity Loss (measured as a percentage) as a function of two key hyperparameters: the watermark strength ($\alpha$) and the editing budget ($k$). As discussed in the main paper, this analysis reveals a highly favorable operational landscape, where a wide region of near-optimal robustness can be achieved with minimal impact on fidelity. Our default operating point ($\alpha=0.1, k=0.1\%$), marked by the star, is strategically chosen to lie within this "sweet spot," demonstrating the practical controllability afforded by our disentangled framework.

\section{Evaluation Metrics}
\label{app:metrics}

This section provides detailed definitions and mathematical formulations for the evaluation metrics used in our experiments. Following established practices in graph watermarking literature \cite{peng2025kgmark}, we evaluate DRGW across three fundamental dimensions: \textbf{detectability}, \textbf{robustness}, and \textbf{transparency}.

\subsection{Detectability and Robustness}

We employ the Area Under the Receiver Operating Characteristic Curve (AUC) as the primary metric for evaluating both detectability and robustness. The ROC curve plots the True Positive Rate (TPR) against the False Positive Rate (FPR) at various classification thresholds, providing a comprehensive view of the watermark detection performance.

\textbf{Area Under Curve (AUC):} The AUC quantifies the overall ability of our watermark detection system to distinguish between watermarked and unwatermarked graphs. Formally, given a set of positive samples (watermarked graphs) and negative samples (unwatermarked graphs), let $N_p$ and $N_n$ denote their respective counts. For each positive sample with detection score $y_i$ and negative sample with score $y_j$, the AUC is computed as:
    \begin{equation}
        \text{AUC} = \frac{1}{N_p N_n} \sum_{i=1}^{N_p} \sum_{j=1}^{N_n} \mathbb{I}(y_i > y_j),
    \end{equation}
    where $\mathbb{I}(\cdot)$ is the indicator function that returns 1 if the condition is true and 0 otherwise. An AUC value of 1.0 indicates perfect detection, while 0.5 represents random guessing.

For robustness evaluation, we compute AUC under various attack scenarios, including structural perturbations (edge flipping, node deletion) and adversarial attacks (NEA, L2 Metric). The robustness AUC measures the watermark's resilience against these modifications.

\subsection{Transparency}

We assess transparency from two complementary perspectives: \textbf{structural fidelity} and \textbf{functional fidelity}.

\subsubsection{Structural Fidelity Metrics}

Structural fidelity evaluates how well the watermarked graph preserves the topological properties of the original graph. We employ the following metrics:

 \textbf{Flipped Edges (\%):} The percentage of edges that are modified (added or removed) during the watermark embedding process relative to the total number of edges in the original graph:
    \begin{equation}
        \text{Flipped Edges (\%)} = \frac{|E \oplus E_w|}{|E|} \times 100\%,
    \end{equation}
    where $E$ and $E_w$ are the edge sets of the original and watermarked graphs, respectively, and $\oplus$ denotes the symmetric difference.
    
\textbf{Assortativity Change (\%):} The relative change in the degree assortativity coefficient, which measures the tendency of nodes to connect with other nodes of similar degree:
    \begin{equation}
        \text{Assortativity Change (\%)} = \left| \frac{r - r_w}{r} \right| \times 100\%,
    \end{equation}
    where $r$ and $r_w$ are the assortativity coefficients of the original and watermarked graphs, respectively.
    
\textbf{Clustering Coefficient Change (\%):} The relative change in the global clustering coefficient, which quantifies the tendency of nodes to form triangles:
    \begin{equation}
        \text{Clustering Coefficient Change (\%)} = \left| \frac{C - C_w}{C} \right| \times 100\%,
    \end{equation}
    where $C$ and $C_w$ are the global clustering coefficients of the original and watermarked graphs, respectively.
    
\textbf{dK-2 Deviation:} A graphlet-based similarity measure that captures fine-grained structural differences between graphs. The dK-2 series measures the distribution of connected degree pairs in the graph, and the deviation is computed as the Earth Mover's Distance between the dK-2 distributions of the original and watermarked graphs.

\subsubsection{Functional Fidelity Metrics}

Functional fidelity evaluates how well the watermarked graph preserves its utility for downstream applications. We employ the following metrics:

\textbf{Link Prediction Performance Drop (\%):} The relative decrease in performance on the link prediction task after watermarking:
    \begin{equation}
        \text{Link Prediction Drop (\%)} = \left(1 - \frac{\text{AUC}_{\text{pred}}^{w}}{\text{AUC}_{\text{pred}}}\right) \times 100\%,
    \end{equation}
    where $\text{AUC}_{\text{pred}}$ and $\text{AUC}_{\text{pred}}^{w}$ are the AUC scores of link prediction on the original and watermarked graphs, respectively.
    
\textbf{Cosine Similarity:} The cosine similarity between the node embeddings learned from the original graph and those learned from the watermarked graph. For two embedding matrices $\mathbf{E}$ and $\mathbf{E}_w$ of the same dimension, the cosine similarity is computed as:
    \begin{equation}
        \text{Cosine Similarity} = \frac{1}{N} \sum_{i=1}^{N} \frac{\mathbf{e}_i \cdot \mathbf{e}_{w,i}}{\|\mathbf{e}_i\| \|\mathbf{e}_{w,i}\|},
    \end{equation}
    where $N$ is the number of nodes, and $\mathbf{e}_i$, $\mathbf{e}_{w,i}$ are the embedding vectors of node $i$ in the original and watermarked graphs, respectively.

For all transparency metrics except Cosine Similarity, lower values indicate better performance (i.e., less distortion). For Cosine Similarity, higher values indicate better preservation of the original graph's embedding structure.

\section{Visual Analysis of Disentanglement}
\label{app:disentanglement_vis}

\begin{figure*}[t!]
  \centering
  \Description{Three t-SNE plots. The first shows entangled representations. The second and third show that DRGW successfully separates structural information from watermark carrier information.}

  \begin{subfigure}[b]{0.32\textwidth}
    \centering
    \includegraphics[width=\linewidth]{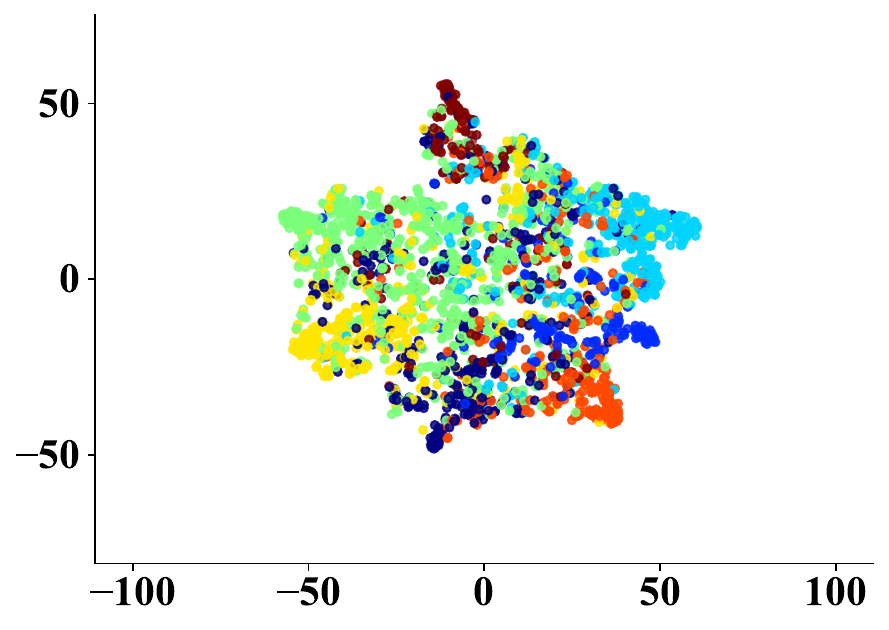}
    \caption{Entangled Subspace}
    \label{fig:tsne_entangled}
  \end{subfigure}
  \hfill
  \begin{subfigure}[b]{0.32\textwidth}
    \centering
    \includegraphics[width=\linewidth]{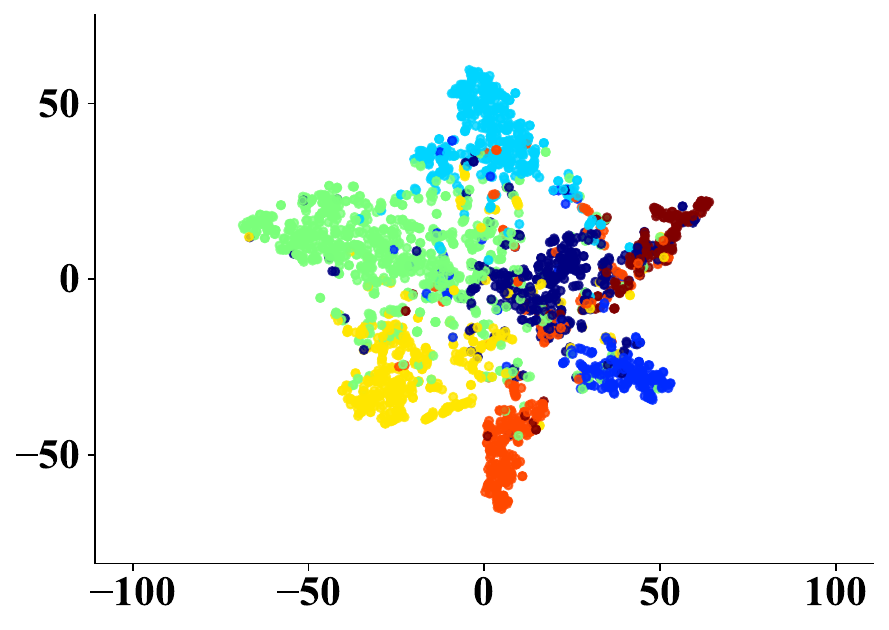}
    \caption{Our Structural Subspace ($h_s$)}
    \label{fig:tsne_ours_hs}
  \end{subfigure}
  \hfill
  \begin{subfigure}[b]{0.32\textwidth}
    \centering
    \includegraphics[width=\linewidth]{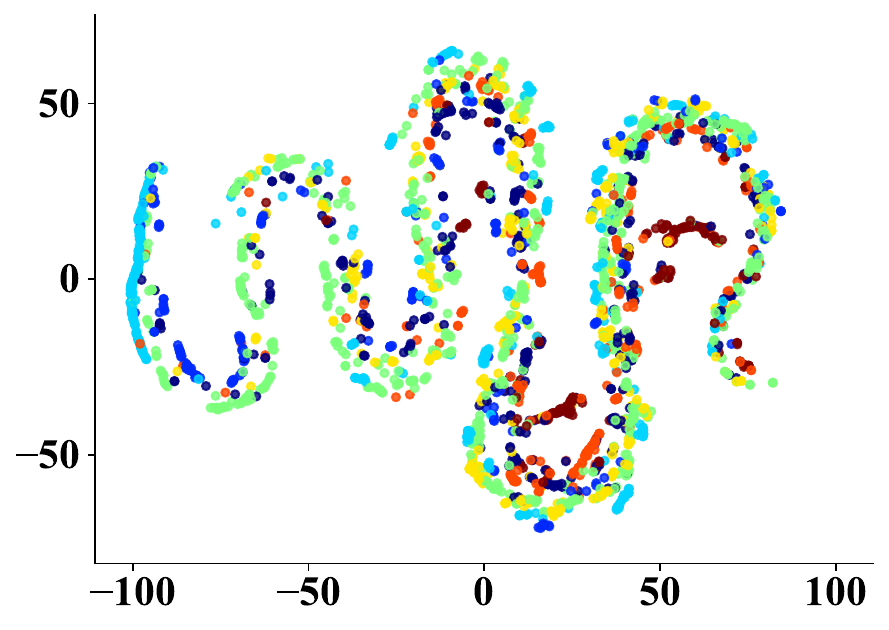}
    \caption{Our Carrier Subspace ($h_w$)}
    \label{fig:tsne_ours_hw}
  \end{subfigure}
  \vspace{-0.2cm}
  \caption{Mechanism Verification: Visual proof of latent space disentanglement. (a) In the entangled subspace, information is mixed. (b, c) DRGW successfully separates structural ($h_s$) and carrier ($h_w$) information, preventing mutual interference.}
  \label{fig:tsne_disentanglement}
\end{figure*}

To explicitly demonstrate the effectiveness of our disentanglement strategy discussed in the Ablation Study, we present the t-SNE visualizations of the latent spaces in Figure~\ref{fig:tsne_disentanglement}.

\section{Visual Analysis of Robustness}
\label{app:robustness_vis}

\begin{figure}[h!]
    \centering
    \Description{Two t-SNE plots of the carrier subspace under attack. The left plot shows the result after 50\% node deletion, and the right plot shows the result after 50\% edge flipping. In both cases, the overall shape of the distribution remains, although it is more diffuse.}
   
    \begin{subfigure}[b]{0.48\columnwidth}
        \centering     
        \includegraphics[width=\textwidth]{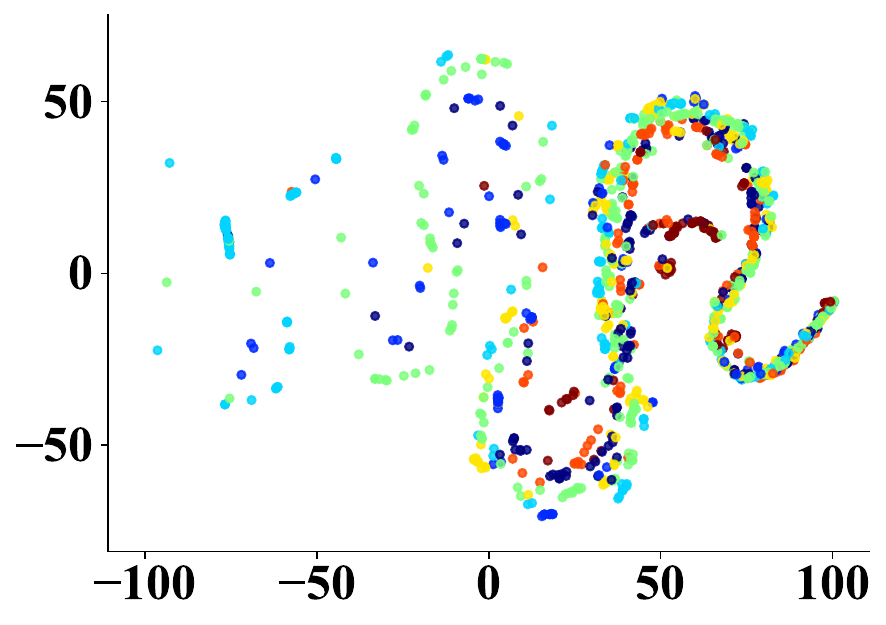}
        \caption{After 50\% Node Deletion}
        \label{fig:robust_node_del}
    \end{subfigure}
    \hfill
    \begin{subfigure}[b]{0.48\columnwidth}
        \centering
        \includegraphics[width=\textwidth]{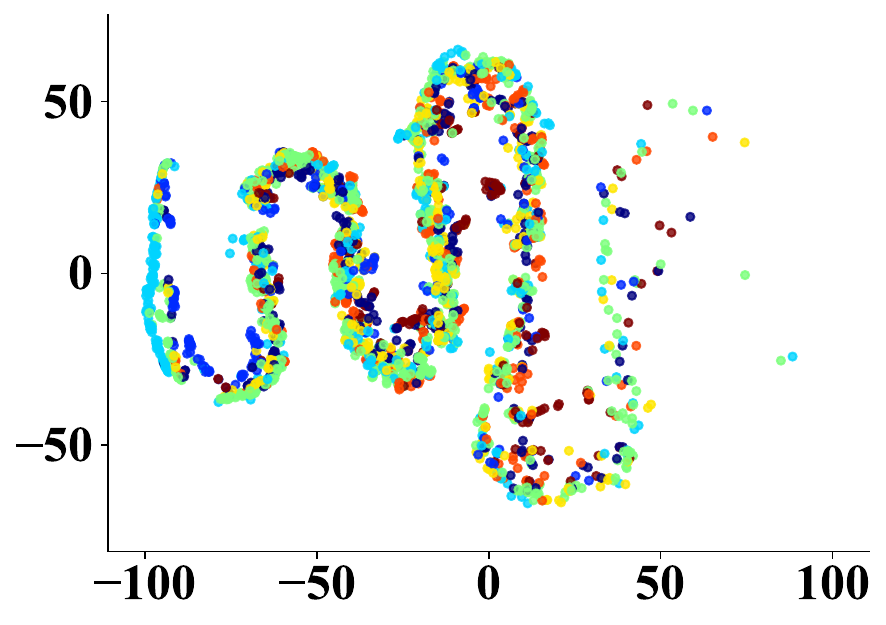}
        \caption{After 50\% Edge Flipping}
        \label{fig:robust_edge_flip}
    \end{subfigure}
    \vspace{-0.2cm}
    \caption{Resilience of the Carrier Subspace $h_w$: The distribution maintains its structural integrity even under severe 50\% node deletion (a) and edge flipping (b) attacks. The macroscopic structure is preserved, enabling effective watermark recovery.}
    \label{fig:robustness_visualization} 
\end{figure}

To visually corroborate the resilience of our framework discussed in Section~\ref{sec:robustness}, we present the t-SNE visualizations of the carrier subspace ($h_w$) under severe structural attacks in Figure~\ref{fig:robustness_visualization}.

\section{Quantitative Analysis of Disentanglement}
\label{app:disentanglement_quant}

To complement the qualitative t-SNE visualizations in Figure~\ref{fig:tsne_disentanglement}, this section provides a quantitative validation of the disentanglement between the structural representation $h_s$ and the watermark carrier $h_w$. We adopt two standard metrics from the representation learning literature: the Mutual Information Gap (MIG)~\cite{chen2018isolating} and DCI (Disentanglement, Completeness, and Informativeness)~\cite{eastwood2018framework}. Specifically, MIG quantifies how concentrated the information about a single ground-truth factor is within a single latent dimension; a higher score indicates better disentanglement. The DCI framework provides two key scores used in our evaluation: DCI-Disentanglement measures if each latent dimension captures at most one factor, penalizing redundancy, while DCI-Informativeness assesses the overall predictive power of the representation for determining these factors.

\subsection{Experimental Setup}
Since real-world graphs lack explicit ground-truth generative factors, we construct a synthetic dataset where a key structural property can be precisely controlled. We generate a set of 2,000 Barabási-Albert graphs, each with 100 nodes. We then systematically vary the global clustering coefficient of these graphs by adding or removing edges involved in triangular motifs, creating 10 discrete bins that serve as our single ground-truth factor of variation.

For each synthetic graph, we use our pre-trained DRGW encoder to extract the representations $h_s$ and $h_w$. To compute the MIG and DCI scores, we train a gradient boosting classifier to predict the binned clustering coefficient from the latent representations.

\subsection{Results}
The results summarized in Table~\ref{tab:disentanglement_appendix}, quantitatively confirm the effectiveness of our disentanglement strategy. The structural representation $h_s$ achieves high scores across all metrics, indicating that it is highly informative about the graph's structural properties and that this information is concentrated in a few latent dimensions. Conversely, the watermark carrier $h_w$ shows negligible predictive power, with scores near zero. This provides strong empirical evidence that our framework successfully isolates structural information within $h_s$, leaving $h_w$ as a clean, independent channel for watermarking. The entangled representation from the Naive Baseline falls in between, demonstrating a partial but incomplete separation of information.

\begin{table}[htbp]
\centering
\caption{Quantitative disentanglement evaluation. Higher scores are better for all metrics. Results confirm that structural information is almost exclusively captured by $h_s$.}
\label{tab:disentanglement_appendix}
\small
\begin{tabular}{l c c c}
\toprule
\textbf{Representation} & \textbf{MIG} & \textbf{DCI-Disent.} & \textbf{DCI-Info.} \\
\midrule
DRGW ($h_s$) & \textbf{0.412} & \textbf{0.853} & \textbf{0.921} \\
DRGW ($h_w$) & 0.015 & 0.124 & 0.088 \\
\midrule
Naive Baseline (Entangled) & 0.189 & 0.467 & 0.905 \\
\bottomrule
\end{tabular}
\end{table}

\end{document}